# Numerical Atrribute Extraction from Clinical Texts

1 AUTHOR:

Sunil Mandhan
Hitachi, Ltd.

**2** PUBLICATIONS    **0** CITATIONS

# Extraction and Association of Numerical Attributes and Values from Electronic Health Records

[1]Sarath P R, [1]Sunil Mandhan, and [2]Yoshiki Niwa

[1]Research and Development Centre, Hitachi India Pvt. Ltd., Bangalore, India
`{sarath,sunilm}@hitachi.co.in`
[2]Hitachi Ltd., Central Research Laboratory, Japan
`yoshiki.niwa.tx@hitachi.com`

**Abstract.** This paper describes about information extraction system, which is an extension of the system developed by team Hitachi for "Disease/Disorder Template filling" task organized by ShARe/CLEF eHealth Evolution Lab 2014. In this extension module we focus on extraction of numerical attributes and values from discharge summary records and associating correct relation between attributes and values. We solve the problem in two steps. First step is extraction of numerical attributes and values, which is developed as a Named Entity Recognition (NER) model using Stanford NLP libraries. Second step is correctly associating the attributes to values, which is developed as a relation extraction module in Apache cTAKES framework. We integrated Stanford NER model as cTAKES pipeline component and used in relation extraction module. Conditional Random Field (CRF) algorithm is used for NER and Support Vector Machines (SVM) for relation extraction. For attribute value relation extraction, we observe 95 % accuracy using NER alone and combined accuracy of 87 % with NER and SVM.

**Keywords:** NLP, NER, relation extraction, information extraction, crf, svm

## 1    Introduction

Healthcare providers are increasingly adopting Electronic Health Record (EHR) systems to improve the quality of care. Nowadays EHR data and systems are accessible to patients (patient portals) and non-expert clinical professionals. Clinical information inside EHR systems are various and mainly are in the form of unstructured text (e.g. Discharge Summary). It is difficult for non-expert end users to interpret the documents which contain many medical abbreviations and jargons [8]. Extracting frequently required information from unstructured clinical text and representing in a structured manner will give quick and timely access to patient's health related data to end users. For example information about patient's body vital signs, blood components, drugs etc. are used in day to day operations to understand the progress and treatment of patient. Most of such information is in numerical form. In this paper we describe about the experiment, developed system and results of numerical attributes and related values extraction from discharge summary records.

## 1.1 Problem Description

Attributes are originated from physical examinations and medical tests required for disease diagnosis as well as treatment procedures. For example, blood pressure and heart rate are the common and important numerical measurements required for diagnosis of almost all the diseases. Table 1 shows some important numerical attributes which are prominent in clinical diagnosis and found in discharge summaries [8].

**Table 1.** Types and examples of numerical attributes

| Class | Examples of attributes |
|---|---|
| Vital Signs | Blood pressure, temperature, pulse, heart rate, respiratory rate, oxygen saturation |
| Blood Components | WBC, RBC, hematocrit, platelets count, glucose, urea, nitrogen, sodium, potassium, anion gap |
| Drug Attributes | Dosage, quantity, frequency, periodic interval |

**Example Sentence.** "Her vital signs the following day, she had heart rate of 66, blood pressure 120/63, respiratory rate 14, 100% on 5 liters nasal cannula O2 saturation."

**Manual Annotation.** "Her vital signs the following day, she had <attribute-1>heart rate<attribute-1> of <value-1>66<value-1>, <attribute-2>blood pressure</attribute-2> <value-2>120/63</value-2>, <attribute-3>respiratory rate</attribute-3> <value-3>14</value-3>, <value-4>100%</value-4> on 5 liters nasal cannula <attribute-4>O2 saturation</attribute-4>."

## 2 Related Work

There are many references available for extracting different types of information from clinical documents. Identifying medications, tests, procedures [2], symptoms, protein names [3], enzyme interactions and protein structures [4] are some examples.
Other kinds of information extraction include summarization of medical documents to a tabular format by identifying the events, time and negativity [9].

I2B2 [5] and CLEF [6] are the notable workshops which are engaged in organizing NLP research challenges in medical domain. The CLEF paper [1] has used SVM for relation extraction for detecting relationship between disease and different properties of disease like body location, severity, etc. SVM was used in case where more than one body locations were present to establish the relationship which body location is associated to the disease.

To the best of our knowledge, we could not find any related work where numerical attributes and values extraction from clinical documents is designed and evaluated as a relation extraction problem using a combination of supervised machine learning techniques. In this paper, we attempt to solve this problem and explain the different algorithms with input features and relevant data pre-processing.

## 3    System Architecture

We solve the problem in two steps. First step is extraction of numerical attributes and values, which is developed as a Named Entity Recognition model using Stanford NLP libraries. Second step is correctly associating those attributes to values, which is developed as a relation extraction module in Apache cTAKES framework. We integrated Stanford NER model as a cTAKES pipeline component and used in relation extractor. Figure 1 describes the architecture of the developed system. Section 3.1 describes about algorithm and approaches in attributes and values extraction. Section 3.2 describes about relation extraction module.

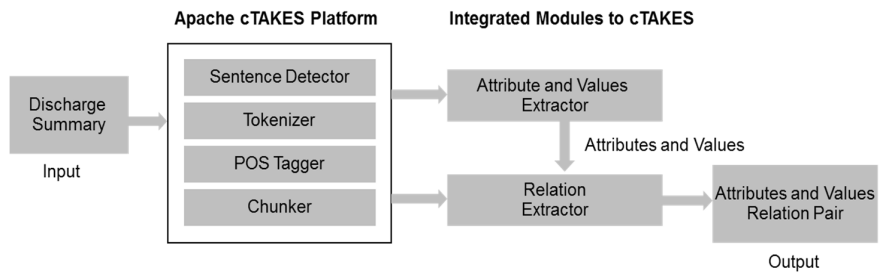

**Fig. 1.** System Architecture

### 3.1    Attributes and Values Extraction

We trained a model for Attribute and Value extraction from discharge summaries using Stanford NER library. Stanford NER provides a general implementation of arbitrary order linear chain Conditional Random Field (CRF) sequence models [10]. It can be trained for any task.

#### 3.1.1    Tokenization:
We observed the default Stanford Tokenizer (Penn Treebank Tokenizer) splits certain attribute words in discharge summary in an undesired manner as explained with an example below.

| Word | Required Tokenization | Output of Stanford Tokenization |
|---|---|---|
| WBC-12.8*# | WBC, - , 12.08, *, # | WBC-12, .8, *, # |

To avoid this issue we applied a regular expression based preprocessing before input data for Tokenization. We replaced the hyphens in certain types of attributes with white space as it is harmless to do so. Following regular expression is used for preprocessing.

Regular expression: ([a-z A-Z]|O3|O2|B12)(-)([0-9])

After preprocessing, we get the following tokenization

| Word | Required Tokenization | Output of Stanford Tokenization |
|------|----------------------|--------------------------------|
| WBC 12.8*# | WBC, 12.08, *, # | WBC,12.8, *, # |

Modification to the regular expression or new implementation of tokenizer is applicable as and when more issues are detected in tokenization.

**3.1.2    Model Training:** From the tokens and our manually annotated discharge summaries, we programmatically prepared training data for Stanford CRF Classifier. We used the following features from Stanford NER Feature Factory [11] for training the CRF model.

1. Word
2. Position (word index in sentence)
3. Word shape feature
4. Ngrams from word
5. Disjunctions of words

## 3.2    Relation Extraction

SVM algorithm is used for establishing the relationship between attribute and value. SVM is a distance based method and has proved to be effective for relation extraction [1]. The basic idea of using SVM on relationships is to map a relation into a feature space and find the maximum margin hyper plane to separate two classes (related and not related).

**3.2.1    Model Training:** We trained SVM model for relation association using manually annotated discharge summaries. Following features are used.

1. Part of speech
2. Punctuation
3. Phrase chunking (Noun phrase, Verb phrase, etc.)
4. Attribute presence feature: This feature is used to check if there is any other attribute present between an attribute and value pair for which relationship is being predicted. Example: 1+ right DP pulse, 2+ left PT pulse In this example, "right DP pulse" attribute is present between 1+ (value) and "left PT pulse" (attribute).
5. Distance feature: This feature captures the distance (number of tokens) between an attribute and value pair. Example: Lactate elevated at 6. In this example, distance between Lactate (attribute) and 6 (value) is 2.

## 4 Results

This work being an extension of the previously developed system for CLEF eHealth 2014 task, we use the same data which was served as training corpus for CLEF eHealth 2014 task 2. In this work we have experimented only discharge summary records from the available data. We split the total available 136 records into 100 and 36 for training and test purpose respectively.

### 4.1 Evaluation Criteria

The matching between detected values and true values in actual data are done in a strict manner. These detected values and true values are actually sequences of characters (i.e. string) in a text. Thus, in the strict evaluation, detected and true values are compared literally.

For example, the true representation for attribute in actual data is "blood pressure"; an identified attribute by the system should be identical to it, i.e. "blood pressure" in order to be marked as true. Other outputs including matches in substrings like "blood" or "pressure" will be marked as false.

### 4.2 Evaluation results

We share the evaluation results of two runs with different features in Table 2. In Run 1 we have used all the features described in section 3.1.2 and first three features described in section 3.2.1. In Run 2 we have used all the features described in sections 3.1.2 and 3.2.1.

**Table 2.** - Strict evaluation of CRF and SVM

| CRF | | | SVM | | |
|---|---|---|---|---|---|
| Accuracy Type | Run1 Value | Run2 Value | Accuracy Type | Run1 Value | Run2 Value |
| Recall | 0.93 | 0.93 | Recall | 0.75 | 0.83 |
| Precision | 0.97 | 0.97 | Precision | 0.92 | 0.93 |
| F-score | 0.95 | 0.95 | F-score | 0.83 | 0.87 |

### 4.3 Discussion

The most important finding during this research is that tokenization and feature engineering is very important in NLP based systems.

Another important finding was the distribution of positive and negative training samples. Training data needs to be explicitly checked for the distribution. During the relation extraction testing, accuracy was very low and was not improving. It was found during debugging of the SVM tool that the training data was extremely biased towards negative data samples. This biasing made the SVM classifier classifying every test data sample to negative category. After the distribution was corrected and biasing was tuned, results were improved drastically.

## 5 Conclusion

This report describes the approach, algorithms, and tools used in building the numerical attribute and values extraction. CRF algorithm was evaluated and suggested for extracting the attribute and values; it gave 0.95 in F-score. SVM algorithm was evaluated and suggested for relation extraction between an attribute and value, and it gave F-score of 0.87. Though the F-score signifies good accuracy figures, still there is scope of further improvement by designing new features, cross validation, and ensembles of methods (e.g., bagging and AdaBoost). This work can be extended to the problem of non-numerical attribute and value extraction (e.g. dosage mentions such as 'small dose', 'sliding scale', etc. and frequency information like 'two weeks' etc.).